\documentclass[wcp]{jmlr}

\usepackage{longtable}

\usepackage{booktabs}

\usepackage{lineno}
\usepackage{graphicx}
\usepackage{amsmath}
\usepackage{booktabs}
\usepackage{multirow}
\usepackage{color}
\counterwithout{figure}{chapter}

\makeatletter
\let\Ginclude@graphics\@org@Ginclude@graphics 
\makeatother

\jmlrvolume{222}
\jmlryear{2023}
\jmlrworkshop{ACML 2023}

\title[A Novel Counterfactual Data Augmentation Method for Aspect-Based Sentiment Analysis]{A Novel Counterfactual Data Augmentation Method for Aspect-Based Sentiment Analysis}

\begin{document}

 \author{
 \Name{Dongming Wu} \Email{wudongming@myhexin.com} \\
 \addr Hithink RoyalFlush Information Network Co.,Ltd, Hangzhou, China \\
  \Name{Lulu Wen} \Email{wenlulu@zju.edu.cn} \\
  \addr College of Computer Science and Technology, ZhejiangUniversity, Hangzhou, China
  \AND
  \addr Hithink RoyalFlush Information Network Co.,Ltd, Hangzhou, China \\
  \Name{Chao Chen} \Email{chenchao6@myhexin.com} \\
  \addr Hithink RoyalFlush Information Network Co.,Ltd, Hangzhou, China \\
  \Name{Zhaoshu Shi} \Email{shizhaoshu@myhexin.com} \\
  \addr Hithink RoyalFlush Information Network Co.,Ltd, Hangzhou, China
  }
\editors{Berrin Yan{\i}ko\u{g}lu and Wray Buntine}

\maketitle

\begin{abstract}
Aspect-based-sentiment-analysis (ABSA) is a fine-grained sentiment evaluation task, which analyzes the emotional polarity of the evaluation aspects. Generally, the emotional polarity of an aspect exists in the corresponding opinion expression, whose diversity has great impact on model’s performance. To mitigate this problem, we propose a novel and simple counterfactual data augmentation method to generate opinion expressions with reversed sentiment polarity. In particular, the integrated gradients are calculated to locate and mask the opinion expression. Then, a prompt combined with the reverse expression polarity is added to the original text, and a Pre-trained language model (PLM), T5, is finally was employed to predict the masks. The experimental results shows the proposed counterfactual data augmentation method performs better than current augmentation methods on three ABSA datasets, i.e. Laptop, Restaurant, and MAMS. 
\end{abstract}

\section{Introduction}
\par\setlength\parindent{1em}Traditional sentiment analysis tasks consider sentences or documents as the object to analyze their sentiment polarity. However, a sentence may contain multiple aspects with different emotional polarities \citep{xu2019bert,song2019attentional,wang2020relational,zhao2020modeling}. This bring much difficulty for the more fine-grained sentiment analysis. 

\par\setlength\parindent{1em}Pre-trained language models (PLMs) that were trained on massive data in an unsupervised manner, possess great natural language understanding ability and have been fine-tuned to solve various downstream tasks effectively, e.g. sentiment analysis, textual entailment, text summarization, question answering, etc. In practice, the samples are generally concatenated with the aspect words as the input of PLMs, and the parameters of PLMs can be freezed or tuned during training \citep{song2019attentional}. Recently, due to the requirement of few computing resources during fine-tuning and good performance in solving downstream tasks, some new fine-tuned methods, e.g. Prefix tuning \citep{li-liang-2021-prefix}, P-tuning \citep{liu2022p}, LoRA \citep{hu2021lora} etc., which add some external structures while freezing the parameters of PLMs, have drawn widespread attention.

\par\setlength\parindent{1em}Aspect-based-sentiment-analysis (ABSA) can be roughly divided into two subtasks: (1) Extracting the evaluation objects from the text. (2) Judgement of the emotional polarity of the objects. Since a sentence might contain multiple aspects and the opinion expressions may exist implicitly in the text, it is hard to identify all the opinion expressions accurately. Meanwhile, the sample diversity in ABSA tasks is usually insufficient, and this further affects the performance of the fine-tuned models.

\par\setlength\parindent{1em}Being the simplest method, data augmentation which helps to improve the diversity of training samples, can thus be used to alleviate the above issues. Data augmentation methods can be roughly divided into two categories: the modification of existing samples and the generation of new samples \citep{DBLP:journals/corr/abs-1911-03118,DBLP:journals/corr/abs-2003-02245}, where the modification methods can be further divided into noising \citep{wei2019eda}, thesauruses \citep{DBLP:journals/corr/ZhangZL15}, machine translation \citep{sennrich2015improving} and language models \citep{wu2019conditional,DBLP:journals/corr/abs-1909-10351}.

\par\setlength\parindent{1em}Due to the ability to consider context semantics and alleviate the ambiguity problem, language models are ideal for fine-grained natural language processing (NLP) tasks. However, language model-based methods have shortcomings in limiting the word level and affecting the sentence semantics if there are excessive random substitutions. Meanwhile, as it is a label-preserving method, the modifications are restricted to the same semantic area.

\par\setlength\parindent{1em}Recently, generative large language models (GLLMs) with billion parameters which are trained on tera-scale tokens can deal with various downstream tasks by one shot, few shot, or even zero shot. The performance of GLLMs is even better than the models fine-tuned on the specific training data of ABSA tasks. However, the high cost of deployment and the strict governmental policy have made the access to GLLMs difficult. Therefore, it is of great significance to develop simple, resource-friendly and effective ABSA method.

 \par\setlength\parindent{1em}This paper proposes a novel counterfactual data augmentation method for ABSA task. The proposed method is a language model-based method and can be mainly divided into two stages. The integrated gradients is first used to identify the opinion words which thereafter will be masked. Next, the prompts with reversed polarity are added to the original sentences and the language model T5 is employed to predict the masks. In this way, the original aspect polarity is reversed while only modifying a few tokens, and therefore, the model can obtain stronger generalization ability. The proposed method was tested on three open datasets, e.g., Restaurant, Laptop and MAMS, and the results show that the proposed method performs better than several common data augmentation methods.

\par\setlength\parindent{1em}The main contributions of this paper are as follows:
\par\setlength\parindent{1em}(1) We proposed a two-stage data augmentation method composed of opinion corruption and diverse opinion generation.
\par\setlength\parindent{1em}(2) The proposed counterfactual data augmentation method is simple and easy to implement in the production environment, and can be combined with other baseline models to further improve the results of ABSA tasks.
\par\setlength\parindent{1em}The reminder of this paper is arranged as follows. Section 2 introduces the related works. Section 3 presents the proposed counterfactual data augmentation method. The experimental results and conclusion are given in Section 4 and Section 5 respectively.

\section{Related Work}

\subsection{ABSA}

\par\setlength\parindent{1em}Since the introduction of BERT \citep{devlin2018bert}, PLMs based on the Transformer architecture \citep{vaswani2017attention} has transformed the paradigm of NLP. These models are now dominating downstream tasks including ABSA tasks owing to their strong natural language understanding capability. For example, BERT-SPC \citep{song2019attentional} leverages PLMs to incorporate both the text and the aspect into the model input, and thus allowing effective modeling of the relationships between aspects and opinion expressions. Furtherly, adversarial training \citep{karimi2021adversarial}, layer aggregation \citep{karimi2020improving} and domain adaption \citep{rietzler2019adapt} were integrated to improve the performance.

\par\setlength\parindent{1em}As a fine-grained NLP task, it is important to incorporate syntactic information into ABSA. This can guide the model to focus on the relevant parts of the aspects, which in turn can help to solve the opinion identification problem more effectively. AdaRNN \citep{dong2014adaptive} transformed the dependency tree, starting from the aspect words, into a recursive structure which was modeled with Recursive Neutral Network (RNN). As dependency information are not entirely accurate and needs to be corrected for accurate target identification, \citet{he-etal-2018-effective} incorporated syntactic information with attention mechanism. \citet{zhang2019syntax} then applied proximity weight on both position and dependency, \citet{wang2020relational} then reshaped and pruned the original dependency tree.  Yet the derived dependency tree can only represent external criteria, conflicting with the knowledge of the fine-tuned PLMs. \citet{dai2021does} proposed a distance-based method which derivied the dependency trees from fine-tuned PLMs.

\par\setlength\parindent{1em}Another challenge is the multi-aspect problem, where features of different aspects would affect each other. \citet{liang2021enhancing,wang2022contrastive} employed contrastive training objects to tackle the challenge. \citet{yang2021improving} designed a local sentiment aggregation method that can facilitate mutual learning among aspects, enabling the discovery of implicit expressions.

\par\setlength\parindent{1em}In addition, with the development of GLLM, auto-regressive models have been applied for ABSA tasks. For example, \citet{mao2021joint} converted ABSA to a text-to-text task, marking the required emotional elements in the original sentence and using it as the target sequence for the generative model to learn the mapping relationship. \citet{scaria2023instructabsa} directly used the instruction-tuning based model and achieved good results.

\subsection{Data Augmentation}
Data augmentation refers to the modification and expansion of the original data, therefore introducing more samples and increasing the diversity of the training set. Modification means to modify the elements in the sentence without destroying the original sentence structure and label. EDA \citep{wei2019eda} is a simple modification method consisting of synonym replacement, random deletion and insertion. CBERT \citep{wu2019conditional} proposed a mask and predict mechanism where the words were masked randomly and then predicted by BERT. BackTranslation \citep{sennrich2015improving} that can get high-quality, richly diverse samples by translating back and forth, also draw widespread attention. However, the flip side of the coin is that modifications are restricted to the same semantics as a label-preserving method. Expansion means creating new data by generation methods and sampling from it. It is a more flexible way and can be designed for different task descriptions. LAMBDA \citep{DBLP:journals/corr/abs-1911-03118} transforms the classification dataset into a seq2seq dataset and fine-tuned it on GPT-2, then generates new sentences with specific labels. However, such process requires extensive computing resources. 

\par\setlength\parindent{1em}There are also some proprietary augmentation methods for ABSA task. \citet{chen2022unsupervised} extended unsupervised data augmentation methods to span-level. \citet{zhang2022towards} introduced datasets from sentiment analysis and merged them using pseudo-labels. Based on supervised attention, \citet{liang2021enhancing} extracted crucial information and utilized it to generate augmented data. \citet{wang2022contrastive} applied contrastive learning to distinguish different polarities, where negative samples are generated by T5 fine-tuned with Prompt Tuning. \citet{hsu2021semantics} introduced a two-stage method composed of selective perturbed masking and label-preserving token replacement. 

\subsection{Interpretability}
Interpretability refers to the extent to which a person can comprehend the rationale behind a decision. There are two mainstreaming solutions for interpretability: attention mechanism \citep{bahdanau2014neural} based methods and saliency methods. Notwithstanding that attention brings transparency to the model by attention weights between two input units, someone believe that it is not sufficiently interpretable. \citet{pruthi2019learning} proposed a deceptive self-attention method which help explain the interaction of information in transformer. \citet{jain2019attention} discovered that the adversarial attention weights lead to the same prediction as the original ones. \citet{serrano2019attention} found that model did not recognize the most important expressions by intermediate representation erasure. To argue this, \citet{wiegreffe2019attention} believed that attention can be used as explanation in some scenarios, and verified through experiments. 

\par\setlength\parindent{1em}Saliency methods can be further divided into erasure-based and gradient-based methods. \citet{zeiler2014visualizing,li2016understanding} pointed out the importance of input units by erasing them at the input level and dimension level. \citet{li2015visualizing} used the absolute value of the gradient to measure the sensitivity of the input to the label in the ABSA task, while \citet{denil2014extraction} used the product of the gradient and the input as a measure. \citet{sundararajan2017axiomatic} first proposed three axioms, namely sensitivity, implementation invariance and completeness.  


\section{Method}

\par\setlength\parindent{1em}This section presents the details of the proposed counterfactual data augment method for ABSA.

\subsection{Framework}

\begin{figure}[h]
    \begin{center}
        \vspace{10pt}
        \includegraphics[width=0.9\textwidth]{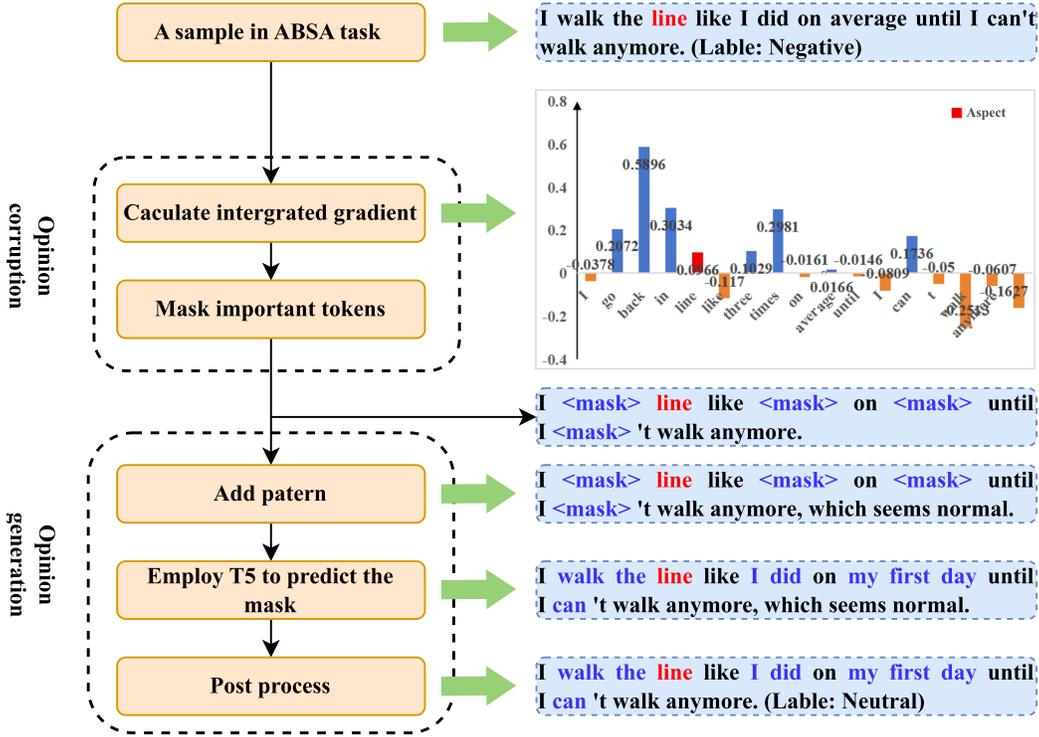}
    \end{center}
    \caption{Framework of the proposed counterfactual data augmentation framework.}\label{fig1}
\end{figure}
 
\par\setlength\parindent{1em}As shown in Fig.\ref {fig1}, the proposed counterfactual data augmentation method for ABSA is mainly composed of two steps, i.e. opinion corruption and opinion generation. The opinion corruption step is designed to identify and mask the most important tokens to the target label. As for the opinion generation step, the designed prompt is added to the masked sentences and the masked tokens will be predicted by T5. Finally, the added prompt is removed from the new sentence and the new label is obtained by a post process.

\subsection{Opinion corruption}
\par\setlength\parindent{1em}To identify the associated opinion words of the given aspect, a quantitative analysis of opinion words is implemented by integrated gradients as described in \citet{sundararajan2017axiomatic}.

\par\setlength\parindent{1em}Assuming the original dataset, training set and test set are $D$, $D_{train}$, $D_{test}$ respectively. $x_{i}$, $A_{i}$, $y_{i}$ represent a training sample, the aspects and the corresponding label respectively. The aspects may contain $k$ aspects $A_{i}=\{a_{i0},...,a_{ik}\}$. 

\par\setlength\parindent{1em}The first step is to train a classifier $M_{base}$ in advance to calculate integrated gradients. Since the existence of data imbalance problem, the balanced cross entropy(BCE) thereby is introduced in the loss function. Suppose in one batch, the numbers of three labels are $n_{0},n_{1},n_{2}$, and the standard cross entropy loss is $L(M(x),y)$, so the BCE can be described by equation \ref{eq6}:

\begin{equation}
    \label{eq6}
    L_{balanced}=\sum_{i=0}^{2}(1-\frac{n_{i}}{\sum_{i}n_{i}}) L(M(x),y)
\end{equation}

\par\setlength\parindent{1em}Then the contribution of each token in $x_{i}=\{r_{0},r_{1},...,a_{0},...,r_{l},[SEP],a_{0}\}$ to the label $y_{i}$ is calculated. $r_{i}$ denotes token, $a_{0}$ denotes aspect. 

\par\setlength\parindent{1em}Specifically, the integrated gradients are the path integral of gradient from a baseline $x_{i0}$ to the input $x_{i}$. Here, we replace all tokens except ones in aspects set with [PAD] as the baseline. Then, the integrated gradients can be represented as equation \ref{eq1}.
\begin{equation}
    \label{eq7}
    x_{i0}=\{[PAD],[PAD],...,a_{0},...,[PAD],[SEP],a_{0}\}
\end{equation}

\begin{equation}
    \label{eq1}
    ig(x_{i})=(x_{i}-x_{i0})\times \int_{\alpha =0}^{1}\frac{\partial M(x_{i0}+\alpha (x_{i}-x_{i0}))}{\partial x_{i}} d\alpha 
\end{equation}

\noindent where $ig(x_{i})$ is the attribution of each token in $x_{i}$ to the label $y_{i}$. 

\par\setlength\parindent{1em}In practice, it is not possible to calculate the consecutive integrals, and thereby we obtain the integrated gradients by a linear interpolation operation. The interpolated sample $\overline{x_{i}}$ is as shown in equation \ref{eq2}.

\begin{equation}
    \label{eq2}
    \overline{x_{i}}=\{x_{i0},x_{i1},...,a_{iS},x_{i}\}
\end{equation}

\begin{equation}
    \label{eq3}
    x_{ij} = x_{i0} + j\frac{x_{i}-x{i0}}{S}
\end{equation}

\noindent where $S$ is the interpolation number.

\par\setlength\parindent{1em}The input $\{\overline{x_{i}}, y_{i}\}$ is then passed into the model $M_{base}$, and a forward and a backward operation are performed to obtain the attribution of each token as shown in following equation.

\begin{equation}
    \label{eq4}
    attr(x_{i})=Norm_{emb}(\sum_{j}grad(x_{ij}))
\end{equation}

\par\setlength\parindent{1em}Here, the large value of $attr(x_{i})$ indicates the higher contribution of $x_{i}$, and vice versa. 

\par\setlength\parindent{1em}Next, the tokens with higher value than the threshold $thr_{con}$ are masked and the continuous masks will be merged, where $thr_{con}=topK(attr(x_{i}, floor(\frac{len(attr(x_{i}))}{3})))$. For instance, in a raw training sentence “Maximum sound isn’t nearly as loud as it should be [SEP] Maximum sound”, the aspect words are “Maximum sound”. The tokens satisfied the threshold are “isn”, “nearly”, “as”, “loud”, “be”, so the masked sample is “Maximum sound $\langle mask \rangle$ ’t $\langle mask \rangle$ as it should $\langle mask \rangle$”.

\subsection{Opinion generation}

\par\setlength\parindent{1em}In opinion generation step, the artificial prompts are first added to the obtained corrupted samples. Compared with the soft prompts, hard prompts methods require no fine-tuning making them more efficient and interpretable.

\par\setlength\parindent{1em}Assuming the masked sample and it's corresponding label are $\Acute{x_{i}}$ and $\Acute{y_{i}}$, the new sample with added prompts and the new label are  $\Tilde{x_{i}}$ and $\Tilde{y_{i}}$ respectively. Consider the above example, the original label negative. Therefore, we should add a positive or neutral prompt, for example “Maximum sound $\langle mask \rangle$ 't $\langle mask \rangle$ as it should $\langle mask \rangle$, which is great!”. 

\par\setlength\parindent{1em}Then, the new sample is passed into T5 model to predict the mask tokens and the generation sample $\overline{x_{i}}=T5(\Tilde{x_{i}})$ is obtained. Moreover, by removing the added prompt $pattern_{i}$ in $\overline{x_{i}}$, we can acquire the final generated sample $\hat{x_{i}}$. In the above example, the final augmented sample is “Maximum sound quality ’thumping’ as it should be”.  

\par\setlength\parindent{1em}However, the sentiment polarities of counterfactual samples may be shifted from $\Acute{y_{i}}$, thereby needing assistance from baseline model $M_{base}$ to determine the final label as follows:

\begin{equation}
    \label{eq5}
    \Tilde{y_{i}}=\begin{cases}
  \Tilde{y_{i}} & \text{ if } argmax(M_{base}(\hat{x_{i}} ))=\Tilde{y_{i}} \text{ and } max(M_{base}(\hat{x_{i}})) > thr_{con}\\
  argmax(M_{base}(\hat{x_{i}} ))& \text{ else } 
\end{cases} 
\end{equation}

\noindent where $thr_{con}$ is the probability threshold. It should note that we add several homogeneous for each reversed label and choose the label with maximum probability fluctuation compared with original sample as the final label. Finally, the augmented data are merged with the original training set.

\section{Experiment}
\subsection{Experimental Settings}
\subsubsection{Datasets}

\par\setlength\parindent{1em}The proposed counterfactual data augmentation method was tested on three common datasets, i.e., SemEval 2014 Restaurant, Laptop \citep{pontiki-etal-2014-semeval} and MAMS \citep{jiang-etal-2019-challenge}. The statistics of the datasets are shown in Table \ref{tab:data_set}. Following previous research \citep{wang2022contrastive, hsu2021semantics}, we adopt accuracy and Macro-F1 as the metrics to evaluate the performance. Note that the original datasets do not include validation set.

\begin{table}[!htb]\centering

    \caption{The statistics of the datasets.}
    \vspace{10pt}
    
        \begin{tabular}{cccccccc}
        \hline
        \multirow{2}{*}{Dataset} & \multicolumn{2}{c}{Positive} & \multicolumn{2}{c}{Neutral} & \multicolumn{2}{c}{Negative}\\ \cline{2-7} 
        \multicolumn{1}{c}{} & \multicolumn{1}{c}{Train} & \multicolumn{1}{c}{Test} & \multicolumn{1}{c}{Train} & \multicolumn{1}{c}{Test} & \multicolumn{1}{c}{Train} & \multicolumn{1}{c}{Test}\\ \hline
        Laptop & 994 & 341 & 870 & 128 & 464 & 169\\
        Restaurant & 2164 & 728 & 807 & 196 & 637 & 196\\ 
        MAMS & 3379 & 400 & 2763 & 329 & 5039 & 607\\ \hline
    
        \end{tabular}
    \label{tab:data_set}
\end{table}

\subsubsection{Comparison experiment settings}

\par\setlength\parindent{1em}First, we compare the proposed counterfactual data augmentation method with other data augmentation methods as shown in the following. 

\begin{itemize}
    \item \textbf{EDA} \citep{wei2019eda}: A simple augmentation method including random insertion, deletion, replacement.
    \item \textbf{BackTranslation} \citep{sennrich2015improving}: Translate the text into another language by machine translation models and then translate it back into the original language. Here, the experimental samples are translated into Chinese which are then translated back to English.
    \item \textbf{C$^{3}$DA} \citep{wang2022contrastive}: A cross-channel data augmentation method aiming to generation negative samples for contrastive learning.
    \item \textbf{Senti-SPM} \citep{hsu2021semantics}: A method composed of selective perturbed masking (SPM) and label-preserving token replacement.
\end{itemize}

\par\setlength\parindent{1em}In addition, the BCE was chosen as the loss function of ABSA to be consistent with C$^{3}$DA, and the foundation model is Roberta-base. 

\par\setlength\parindent{1em}Second, the results obtained by different baseline methods of ABSA, i.e., ASGCN 
 \citep{sun2019aspect}, PWCN \citep{zhang2019syntax}, RGAT \citep{wang2020relational}, SPC \citep{song2019attentional}, MLP \citep{dai2021does}, AEN \citep{song2019attentional}, LCF \citep{DBLP:journals/corr/abs-1912-07976}, were compared when using different foundation models, e.g., Bert, Roberta and  glove.840B.300d. Here, the general cross entropy is selected as the loss function of ABSA to make a fair comparison.

\subsubsection{Hyper Parameters}
\par\setlength\parindent{1em}Following previous studies, the batch size is set as 32, the learning rate is 2e-5 for Bert and Roberta and 1e-3 for glove.840B.300d. Three random seeds was used in all the experiments and we present the average results.

\subsection{Augmentation Comparison Result}

To evaluate the effectiveness of the proposed counterfactual data augmentation method, we compared it with other data augmentation methods and the results are presented in Table \ref{tab:aug_comparison}.

\begin{table}[!htb]\centering
    \caption{Experimental results of ABSA tasks using different data augmentation methods.}
    \scalebox{1.0}{
    \begin{tabular}{cccccccc}
    \hline
    \multirow{2}{*}{Method} & \multicolumn{2}{c}{Laptop} & \multicolumn{2}{c}{Restaurant} & \multicolumn{2}{c}{MAMS}\\ \cline{2-7} 
    \multicolumn{1}{c}{} & \multicolumn{1}{c}{Acc} & \multicolumn{1}{c}{F1} & \multicolumn{1}{c}{Acc} & \multicolumn{1}{c}{F1} & \multicolumn{1}{c}{Acc} & \multicolumn{1}{c}{F1}\\ \hline
    RoBERTa-SPC & 76.91 & 70.47 & 84.73 & 76.15 & 83.61 & 82.88 \\ 
    C$^{3}$DA & 81.83 & 78.46 & 87.11 & 81.63 & - & -\\ 
    EDA & 83.07 & 80.22 & 88.21 & 83.04 & 84.95 & 84.49\\ 
    BackTranslation & 82.6 & 79.13 & 88.12 & 82.36 & 84.73 & 84.45\\ 
    Senti-SPM \& Seq2Seq & 83.7 & 80.82 & 88.39 & 83.05 & - & -\\ 
    Counterfactual & \textbf{83.86} & \textbf{81.39} & \textbf{89.2} & \textbf{84.14} & \textbf{85.33} & \textbf{84.87}\\ \hline

    \end{tabular}
    }
    \label{tab:aug_comparison}
\end{table}

\par\setlength\parindent{1em}From Table \ref{tab:aug_comparison}, it can be seen that data augmentation based methods can obviously improve the accuracy and F1 score on Laptop and Restaurant datasets compared with the baseline method RoBERTa-SPC, while obtain slight improvement on MAMS datasets that contain adequate training samples. EDA outperforms BackTranslation on all three datasets. This may be because EDA causes less damage to the original text as EDA employs random replacement, deletion and insertion to introduce diverse opinion expressions, while BackTranslation may destroy the relationship between emotional expressions and aspect words.

\par\setlength\parindent{1em}It is also can be seen that Senti-SPM\&Seq2Seq achieve the best results except our proposed counterfactual data augmentation method. However, Senti-SPM\&Seq2Seq just replace a few unimportant tokens ignoring some important tokens that containing rich semantics.

\par\setlength\parindent{1em}As for our proposed counterfactual data augmentation method, it achieves the best results compared with the other four data augmentation methods. In addition, the proposed method performs good robustness and generalization as the counterfactual operation enrich the diversity of the original samples.

\par\setlength\parindent{1em}Furthermore, the effects of different foundation models, i.e., BERT, RoBERTa and Static embedings, and different baseline methods, i.e., ASGCN, PWCN, RGAT, SPC, AEN, MLP and LCF, were evaluated and the results are shown in Table \ref{tab:all_comparison}.

\begin{table}[!htb]\centering
    \caption{Results of the proposed counterfactual data augmentation method combined with different foundation models and baseline methods.}
    \scalebox{1.0}{
    \begin{tabular}{cccccc}
    \hline
    \multicolumn{2}{c}{\multirow{2}{*}{Models}} & \multicolumn{2}{c}{Restaurant} & \multicolumn{2}{c}{Laptop}\\ \cline{3-6} 
    \multicolumn{1}{c}{} & \multicolumn{1}{c}{} & \multicolumn{1}{c}{Acc} & \multicolumn{1}{c}{F1} & \multicolumn{1}{c}{Acc} & \multicolumn{1}{c}{F1}\\ \hline
    \multirow{6}{*}{Static Embeddings} & ASGCN & 80.09 & 71.01 & 74.27 & \textbf{69.98}\\ 
    \multicolumn{1}{c}{} & \multicolumn{1}{c}{+ Counterfactual} & \textbf{80.77} & \textbf{71.08} & \textbf{74.33} & 69.72\\ 
    \multicolumn{1}{c}{} & PWCN & 81.10 & 72.29 & 75.34 & 71.21\\ 
    \multicolumn{1}{c}{} & \multicolumn{1}{c}{+ Counterfactual} & \textbf{82.26} & \textbf{74.16} & \textbf{76.02} & \textbf{71.86}\\ 
    \multicolumn{1}{c}{} & RGAT & 81.19 & 71.7 & 73.09 & 68.37\\ 
    \multicolumn{1}{c}{} & \multicolumn{1}{c}{+ Counterfactual} & \textbf{82.00} & \textbf{73.16} & \textbf{74.61} & \textbf{69.64}\\ \hline

    \multirow{10}{*}{BERT} & BERT-SPC & 84.02 & 74.94 & 76.42 & 72.06\\ 
    \multicolumn{1}{c}{} & \multicolumn{1}{c}{+ Counterfactual} & \textbf{84.73} & \textbf{78.03} & \textbf{77.59} & \textbf{72.49}\\ 
    \multicolumn{1}{c}{} & RGAT-BERT & 85.03 & \textbf{78.01} & 78.69 & 74.31\\ 
    \multicolumn{1}{c}{} & \multicolumn{1}{c}{+ Counterfactual} & \textbf{85.21} & 77.77 & \textbf{79.05} & \textbf{75.36}\\ 
    \multicolumn{1}{c}{} & BERT-MLP & 84.33 & 77.22 & 77.77 & 73.39\\ 
    \multicolumn{1}{c}{} & \multicolumn{1}{c}{+ Counterfactual} & \textbf{84.51} & \textbf{77.59} & \textbf{78.14} & \textbf{73.96}\\ 
    \multicolumn{1}{c}{} & AEN-BERT & 81.30 & 71.13 & 77.43 & 72.02\\ 
    \multicolumn{1}{c}{} & \multicolumn{1}{c}{+ Counterfactual} & \textbf{81.30} & \textbf{71.73} & \textbf{77.90} & \textbf{72.78}\\ 
    \multicolumn{1}{c}{} & LCF-BERT & 85.27 & 78.81 & 77.83 & 73.23\\ 
    \multicolumn{1}{c}{} & \multicolumn{1}{c}{+ Counterfactual} & \textbf{85.40} & \textbf{79.69} & \textbf{78.53} & \textbf{74.59}\\ \hline

    \multirow{6}{*}{RoBERTa} & RoBERTa-SPC & 84.73 & 76.15 & 76.91 & 70.47\\ 
    \multicolumn{1}{c}{} & \multicolumn{1}{c}{+ Counterfactual} & \textbf{86.67} & \textbf{79.73} & \textbf{77.79} & \textbf{72.93}\\ 
    \multicolumn{1}{c}{} & RGAT-RoBERTa & 85.3 & 77.75 & 77.64 & 73.9\\ 
    \multicolumn{1}{c}{} & \multicolumn{1}{c}{+ Counterfactual} & \textbf{86.07} & \textbf{79.63}& \textbf{78.68} & \textbf{74.87}\\ 
    \multicolumn{1}{c}{} & RoBERTa-MLP & 86.79 & 79.86 & 83.86 & 80.41\\ 
    \multicolumn{1}{c}{} & \multicolumn{1}{c}{+ Counterfactual} & \textbf{86.83} & \textbf{81.03} & \textbf{83.86} & \textbf{80.61}\\ \hline
    
    \end{tabular}
    }
    \label{tab:all_comparison}
\end{table}

\par\setlength\parindent{1em}It can be seen from Table \ref{tab:all_comparison} that whatever the foundation models, the proposed counterfactual data augmentation method can improve the accuracy and F1 score compared with baseline methods. Since RoBERTa model was trained on larger scale data and can obtain better text embedding, combined the proposed counterfactual data augmentation method with baseline methods acquire the best results. Meanwhile, the results also show that the proposed counterfactual data augmentation method can be easily combined with baseline methods in production.

\subsection{Ablation Study}

\par\setlength\parindent{1em}In this section, we further investigate the impact of different masking and prompting strategies on the results as shown in Table \ref{tab:ablation_study}. 

\begin{itemize}
    \item \textbf{Ramdom-Mask}: Replace the integrated gradients based mask strategy with random mask.  
    \item \textbf{Label-Preserve}: Add prompt with the same polarity of the original sample during opinion generation. 
\end{itemize}

\begin{table}[!htb]\centering
    \caption{Results of different mask strategy and prompting method.}
    \scalebox{1.0}{
    \begin{tabular}{cccccccc}
    \hline
    \multirow{2}{*}{Method} & \multicolumn{2}{c}{Laptop} & \multicolumn{2}{c}{Restaurant} & \multicolumn{2}{c}{MAMS}\\ \cline{2-7} 
    \multicolumn{1}{c}{} & \multicolumn{1}{c}{Acc} & \multicolumn{1}{c}{F1} & \multicolumn{1}{c}{Acc} & \multicolumn{1}{c}{F1} & \multicolumn{1}{c}{Acc} & \multicolumn{1}{c}{F1}\\ \hline
    RoBERTa-SPC & 76.91 & 70.47 & 84.73 & 76.15 & 83.61 & 82.88 \\ 
    Random-Mask & 82.29 & 79.1 & 88.04 & 82.58 & 85.18 & 84.85\\ 
    Label-Preserve & 83.7 & 80.82 & 88.39 & 83.05 & 84.96 & 84.43\\ 
    Counterfactual & \textbf{83.86} & \textbf{81.39} & \textbf{89.2} & \textbf{84.14} & \textbf{85.33} & \textbf{84.87}\\ \hline

    \end{tabular}
    }
    \label{tab:ablation_study}
\end{table}

\par\setlength\parindent{1em}From Table \ref{tab:ablation_study}, it can be seen that the proposed counterfactual data augmentation method which employ an integrated gradient-based mask strategy, performs better than random mask strategy. It is because that random mask may create samples that retain the same emotional semantics, and thus not increase the semantic diversity significantly. Since label-preserve prompting is likely to generate synonyms of original tokens, the semantic diversity of samples is not enriched. Therefore, it can be found that counterfactual prompting method obtain better accuracy and F1 score on all three experimental datasets.

\subsection{Augmented sample analysis}

\par\setlength\parindent{1em}Here, an analysis of augmented samples by different methods is presented to provide a more comprehensive comparison as shown in Table 5.

\par\setlength\parindent{1em}From Table \ref{tab:case_study}, it can be seen that EDA will randomly replaced or deleted some tokens, may leading the change of emotional semantics. BackTranslation maintain the semantics of original sample, but may modify the aspect words. Random-Mask just mask some tokens randomly and thus fail to identify opinion expressions. However, the proposed counterfactual data augmentation method make the key opinion expression changed while modify the emotional semantics, and thus increase the diversity of samples.

\begin{table}[!htb]\centering
    \caption{Case study. The two cases come from the Restaurant and MAMS datasets respectively. The first one is only a single aspect word, and the second one has three aspect words, which are enhanced based on "portions".}
    \scalebox{1.0}{
    \begin{tabular}{p{2.5cm}p{11cm}}
    \hline
    \multicolumn{1}{c}{Methods} & \multicolumn{1}{c}{Examples} \\ \hline
    \multicolumn{1}{c}{Source} & I go back in \textbf{line} like three times on average until I can't walk anymore.\\
    \multicolumn{1}{c}{EDA} & like go back in \textbf{line} i three times on average until i cant walk anymore. \\
    \multicolumn{1}{c}{BackTranslation} & I have to requeue an average of 3 times until I can no longer walk.\\
    \multicolumn{1}{c}{Random-Mask} & I go ( in \textbf{line} like three times) on average until I can 't walk anymore.\\
    \multicolumn{1}{c}{Counterfactual} & I walk the \textbf{line} like I did on my first day until I can 't walk again.\\ \hline

    \multicolumn{1}{c}{Source} & The food is right out of heaven, arrive hungry because the \textbf{portions} are huge but not the prices.\\
    \multicolumn{1}{c}{EDA} & the food is right out of heaven arrive hungry come because the \textbf{portions} are huge merely but not the prices \\
    \multicolumn{1}{c}{BackTranslation} & The food was heaven, arrived hungry as the \textbf{portions} were huge but not overpriced.\\
    \multicolumn{1}{c}{Random-Mask} & The food were so right out of heaven, arrive hungry because are \textbf{portions} is just too huge but not the prices.\\
    \multicolumn{1}{c}{Counterfactual} & the food was out of order when we arrive, \textbf{portions} were small but not at reasonable food and prices.\\ \hline
    
    \end{tabular}
    }
    \label{tab:case_study}
\end{table}

\section{Conclusion}
\par\setlength\parindent{1em}This paper proposed a novel and simple counterfactual data augmentation method for ABSA. An integrated gradient-based method is used to identify key opinion expressions which are masked and then will be predicted by T5 to obtain rich opinion expressions. The experiments show that the proposed counterfactual data augmentation method is superior to other augmentation methods, and achieved good results on public datasets. It is expected that the proposed counterfactual data augment method will have the opportunity to expand to other fine-grained NLP tasks in the future.

\bibliographystyle{IEEEtran}
\bibliography{hyreference}

\end{document}